\pdfoutput=1

\documentclass[11pt]{article}

\usepackage{emnlp2021}

\usepackage{times}
\usepackage{latexsym}
\usepackage{spverbatim}
\usepackage{fancyvrb}
\usepackage{listings}

\usepackage[T1]{fontenc}

\usepackage[utf8]{inputenc}

\usepackage{microtype}

\usepackage{times}  
\usepackage{helvet} 
\usepackage{courier}  
\usepackage{graphicx} 
\usepackage{multirow}
\usepackage{subcaption}
\usepackage{courier}
\usepackage{comment}

\title{TWEETSUMM - A Dialog Summarization Dataset for Customer Service}

\author{Guy Feigenblat\thanks{~~With equal contribution}  \thanks{~~Current address:   guy@piiano.com}, Chulaka Gunasekara\footnotemark[1], Benjamin Sznajder\footnotemark[1], Ranit Aaronov, \\ {\bf David Konopnicki, Sachindra Joshi}
 \\
        IBM Research AI\\
\texttt{\{guyf@il, chulaka.gunasekara@, benjams@il\}.ibm.com}\\ \texttt{\{ranit.aharonov2@, davidko@il\thanks{~~Current address:   david.konopnicki@booking.com}, jsachind@in\}.ibm.com}}

\newcommand{\datasetName}{{\sc TweetSumm}} 

\renewcommand{\UrlFont}{\ttfamily\scriptsize}

\begin{document}

\maketitle

\begin{abstract}

In a typical customer service chat scenario, customers contact a support center to ask for help or raise complaints, and human agents try to solve the issues. In most cases, at the end of the conversation, agents are asked to write a short summary emphasizing the problem and the proposed solution, usually for the benefit of other agents that may have to deal with the same customer or issue. The goal of the present article is advancing the automation of this task. We introduce the first large scale, high quality, customer care dialog summarization dataset with close to 6500 human annotated summaries. The data is based on real-world customer support dialogs and includes both extractive and abstractive summaries. We also introduce a new unsupervised, extractive summarization method specific to dialogs. 


\end{abstract}

\section{Introduction}

Text summarization is the task of creating a short version of a long text, retaining the most important or relevant information. 
In NLP, there are two types of summarization tasks- (1) Extractive summarization, in which segments from the original text are selected to form a summary and (2) Abstractive summarization, in which new natural language expressions are generated for summarizing the text. 
The past 
few years have witnessed a tremendous progress in creating both kinds of summaries using seq2seq models. However, these works have largely focused on  documents such as news and scientific publications ~\cite{Lin2019AbstractiveSA}.

In this paper, we focus on summarizing conversational data between customers and human support agents. 
In many enterprises, once an agent is done with handling a customer request, she is required to create a short summary of the conversation for record keeping purposes. At times, an ongoing conversation may also need to be transferred to another agent or escalated to a supervisor. This also requires creating a short summary of the conversation so far, as to provide the right context to the next handling agent. 

Our main contribution is the release of \datasetName{}, a dataset focused on summarization of dialogs, which represents the rich domain of Twitter customer care conversations \footnote{\url{https://github.com/guyfe/Tweetsumm}}.
The dataset contains close to $6500$ extractive and abstractive summaries generated by human annotators from $1100$ dialogs.
This is the first dataset released to the research community, which focuses on real dialogs, as opposed to previous works focusing on meeting conversations~\cite{McCowan2005AMI}, general {\it chitchat} summarization~\cite{gliwa2019samsum}, or topic descriptions of interviews \cite{zhu2021mediasum}.
Furthermore, the fact that each dialog was annotated by $3$ different crowd-workers, resulting in 
an overall of $6$ summaries for each dialog, provides diversity of summaries.
We performed quality control and assessment to remove erroneous summaries, and to ensure that the collected \datasetName{} summaries are of a high quality.
We evaluate several summarization baselines and further provide a novel unsupervised extractive summarization algorithm, referred to as \textit{NRP   Summ} which outperforms other unsupervised baselines for extractive summarization.
Figure~\ref{figure:summary} shows an example of a \datasetName{} dialog along with a human-generated abstractive summary and two machine-generated summaries - abstractive and extractive summaries.
We propose that the dataset quality and scale, is suitable for developing future models for the dialog summarization task. We hope that releasing \datasetName{} for the community will foster further research. 





\begingroup
\renewcommand{\arraystretch}{0.7} 
\begin{figure}[t]
\centering
\resizebox{0.47\textwidth}{!}{%
\begin{tabular}{p{0.7cm}p{6.7cm}}
\hline
\multicolumn{2}{l}{\tiny{\textit{Original dialog}}} \\
\hline
\tiny{\textbf{Customer}} & \tiny{@Company flight1234 from Miami to LaGuardia smells awful. We just boarded. It’s really really bad.}  \\
\tiny{\textbf{Agent}}  & \tiny{@Customer\_id Allie, I am very sorry about this. Please reach out to a flight attendant to address the odor in the aircraft. *TBW }  \\
\tiny{\textbf{Customer}}  & \tiny{@Company They’re saying it game in from the last flight. They have sprayed and there’s nothing else they can do. It’s gross!}  \\
\tiny{\textbf{Agent}}   &  \tiny{@Customer\_id I'm very sorry about the discomfort this has caused you for your flight! *TBW}  \\
\tiny{\textbf{Customer}}  &  \tiny{@Company It’s not just me! Every person getting on the flight is complaining. The smell is horrific.} \\
\tiny{\textbf{Agent}}   &   \tiny{@Customer\_id Oh no, Allie. That's not what we want to hear. Please seek for one of our crew members on duty for further immediate assistance regarding this issue. Please accept our sincere apologies.  *AOS}
 \\
\tiny{\textbf{Customer}} & \tiny{@Company They’ve brought maintenance aboard. Not a great first class experience :(}
  \\
\tiny{\textbf{Agent}}  &  \tiny{@Customer\_id We are genuinely sorry to hear about your disappointment, Allie. Hopefully, our maintenance crew can fix the issue very soon. Once again please accept our sincere apologies for this terrible incident.  *AOS}
  \\
\tiny{\textbf{Customer}}  &  \tiny{@Company Appreciate it. Thank you!}\\
\tiny{\textbf{Agent}}   &   \tiny{@Customer\_id You are most welcome, Allie. Thanks for tweeting us today.  *AOS}\\
\tiny{\textbf{Customer}} & \tiny{@Company They told us to rebook, then told us the original flight was still departing. We got put back on 1234 but are now in the 1st row instead of the 3rd. Can you get us back in seats 3C and 3D?}\\
\tiny{\textbf{Customer}} &  \tiny{@Company My boyfriend is 6feet tall and can’t sit comfortably at the bulkhead.}\\
\tiny{\textbf{Agent}} & \tiny{@Customer\_id Unfortunately, our First Class Cabin is full on our 1234 flight for today, Allie. You may seek further assistance by reaching out to one of our in-flight crew members on duty.  *AOS }\\
\hline
\multicolumn{2}{l}{\tiny{\textit{Ground truth (human) abstractive summary}}} \\
\hline
& \tiny{Customer complains about smell in flight. Agent updated the customer to seek further assistance by reaching out to one of their in-flight crew members on duty.}\\
\hline
\multicolumn{2}{l}{\tiny{\textit{Automated abstractive summary}}} \\
\hline
&\tiny{Customer is complaining about bad smell in his flight. Agent informed to contact in-flight crew member on duty for further assistance.}\\

\hline
\multicolumn{2}{l}{\tiny{\textit{Automated extractive summary}}} \\
\hline
\tiny{\textbf{Customer}} &  \tiny{Flight1234 from Miami to LaGuardia smells awful.They told us to rebook, then told us the original flight was still departing.}\\
\tiny{\textbf{Agent}} & \tiny{
Unfortunately, our First Class Cabin is full on our 1234 flight for today, Allie.
You may seek further assistance by reaching out to one of our in-flight crew members on duty.}\\

\end{tabular}
}
\caption{\datasetName{} dialog and its summaries}
\label{figure:summary}
\end{figure}
\endgroup

\section{\datasetName{} Dataset}\label{sec:dataset}

\datasetName{} comprises of $1100$ dialogs reconstructed from Tweets that appear in the \textit{Kaggle Customer Support On Twitter} dataset\footnote{\tiny{\url{www.kaggle.com/thoughtvector/customer-support-on-twitter}}}, each accompanied by $3$ extractive and $3$ abstractive summaries generated by human annotators.
The Kaggle dataset, is a large scale dataset based on conversations between consumers and customer support agents on \textit{Twitter.com} \cite{hardalov2018towards}. It covers a wide range of topics and services provided by various companies, from airlines to retail, gaming, music etc. 
Thus, \datasetName{} can serve as a dataset for training and evaluating summarization models for a wide range of dialog scenarios.



For creating the $1100$ dialogs of \datasetName{}
, we 
first 
reconstructed 49,155 unique dialogs from the \textit{Kaggle Customer Support On Twitter} dataset (see section \ref{sec:reconstruct}). Second, we filtered short and long dialogs, containing less than $6$ or more than $20$ utterances, in order to focus on dialogs that are representative of most cases. 
This resulted in 45,547 dialogs with an average length of $22$ sentences\footnote{\tiny An utterance, sometimes termed turn, usually contains more than one sentence.}. Next, in order to represent the customer service scenario, in which a single customer interacts with a single agent, dialogs with more than two speakers were removed. From the remaining 32,081 dialogs, we randomly sampled $1100$ dialogs. These dialogs were sent for generation of summaries using crowd-sourcing on the \textit{Appen.com} platform, as described below.  

\subsection{Dialog Reconstruction Method}\label{sec:reconstruct} The data is delivered via a CSV file where each record contains the following fields: $text$~- the anonymized text of the Tweet, $tweet\_id$~- unique anonymized Tweet ID,  $author\_id$~- unique anonymized author ID, $inbound$~- whether the Tweet is to or from a company, $response\_tweet\_id$~- IDs of Tweets that are responses to this Tweet, $in\_response\_to\_tweet\_id$~- ID of the Tweet this Tweet is in response to, and $created\_at$~- date and time the Tweet was sent. 

In order to reconstruct dialogs from Tweets, we traversed the CSV data recursively using the $in\_response\_to\_tweet\_id$ field. At the end of this process, each dialog is a sorted list of Tweets and their metadata fields. In case several Tweets are posted as response to the same Tweet, they are sorted by their $created\_at$ timesamp. This often happens when a message exceeds the length limit for a single Tweet, and has to be split.



\subsection{Summaries Generation} 

Each annotator was asked to generate one extractive and one abstractive summary for a single dialog at a time. 
When generating the extractive summary, the annotators were instructed to highlight the most salient sentences in the dialog. For the abstractive summaries, they were instructed to write a summary that contains one sentence summarizing what the customer conveyed and a second sentence summarizing what the agent responded. See the supplementary material for a detailed description of the instructions provided to annotators before starting the task.
We collected 3 annotations per dialog, such that overall we obtained $\approx{6600}$ summaries: $\approx{3300}$ extractive summaries, termed hereafter the \textbf{extractive dataset} and $\approx{3300}$ abstractive summaries, termed hereafter the \textbf{abstractive dataset}. As explained in the next section, some summaries were discarded following quality control, and for some dialogs, a second round of summaries collection was done. Overall, \datasetName{} contains $3056$ extractive and $3327$ abstractive summaries.

\subsection{Quality Control and Assessment}\label{sec:quality}

\subsubsection{Quality Control}
To guarantee a high quality level of annotations, multiple measures were taken in advance. We only recruited as crowd-workers, members of an Expert Business Partner channel, who are fluent English speakers. Before an annotator was approved for the task, he or she had to pass a quality control test by annotating 10 dialogs with an acceptable high quality. The quality of those summaries was checked manually. Out of 25 annotators who participated in the test only 10 were approved for the task. 

Following completion of the task, several heuristics were applied to identify and discard bad extractive summaries, and statistics were kept on annotators to identify those, if any, that produced erroneous summaries with high frequency. The applied heuristics included removing summaries containing only one sentence, summaries containing only one side (Customer-only or Agent-only), or summaries starting by an Agent turn. We remove summaries starting by an agent turn since tweeter dialogs begin by a customer raising an issue, and hence the summary is expected to begin with a customer turn.
By these cleansing steps, we removed from our dataset 286 extractive summaries. None of the annotators exhibited a high frequency of such bad summaries, supporting the assumption that these errors are due to technical annotation problems, such as erroneously pressing submit prematurely, rather than an annotator performing poorly on the task in general. 

To further assure the quality of the summaries, we computed on each document and for each annotator the percentage of his selected sentences which were also selected by one of the other annotators. A classical Jacquard score would result in irrelevant low-scores if one of the other annotators selected a large number of sentences, and, thus, we used a slightly adapted version $J$=$|A \cap B| / |A|$ which \emph{punishes} A if he selected a less concise summary. No annotator got an extreme low score and the average scores of the annotators range from $50\%$ to $68\%$. For extra safety, we manually checked the summaries with low $J$ scores and found that they do not appear to be unequivocally erroneous. Rather, the difference in the selection of the sentences was due to similar sentences in the original dialog and to the inherent subjectivity of the task, which is also consistent with previous research \cite{daume2005bayesian}

In addition, we looked for cases where annotators used a repeating, or closely-repeating, text for abstrative summaries of different dialogs. We have identified only 9 such abstractive summaries, which were discarded from the dataset.

\subsubsection{Quality Assessment}
We also used annotators to assess the quality of the summaries generated for \datasetName{}. To achieve a high quality standard we recruited NLP experts instead of using the same pool of crowd-workers that worked on the summaries generation task. 
The annotators were instructed to read the dialog carefully and to select a rating between 1 (lowest score) to 5 (highest score) as an answer to three questions focusing on summary \textit{Coverage} and \textit{Readability}. To this end, $100$ pairs of extractive and abstractive summaries from different dialogs were randomly sampled from \datasetName{}, with 3 experts working on each summary. The obtained median score for all 3 questions is $4$, with average ratings ranging between $3.96$-$4.22$. The questions that were asked along with their average scores and $std$, are described in Table~\ref{table:datasetQuality}. 
In order to evaluate the reliability of this assessment, we followed the approach suggested by ~\cite{toledo2019automatic} to measure agreement between the $3$ annotators over ordinal ratings, by reporting average Kappa values among the possible combinations of two annotators. For the extractive and abstractive Coverage questions, the obtained Kappa scores are 0.41 and 0.56 respectively. For the abstractive Readability question the obtained Kappa score is 0.36. While not perfect, the obtained Kappa values are expected due to the inherent subjectivity of the summarization task, as backed up by previous research~\cite{daume2005bayesian}.

\textbf{We thus conclude, based on our quality control and assessment, that the \datasetName{} dataset contains high quality summaries generated by high quality annotators.}

\begin{table}[]
\resizebox{0.45\textwidth}{!}{%
\begin{tabular}{cll}
\hline
\textit{\textbf{\begin{tabular}[c]{@{}c@{}}Summary\\ Type\end{tabular}}} & \textit{\textbf{Question}} & \textit{\textbf{\begin{tabular}[c]{@{}l@{}}Average\\ score\end{tabular}}} \\ \hline
\textit{extractive} & \begin{tabular}[c]{@{}l@{}}Provide your rating as to the overall  \textbf{coverage} of\\ the summary, based on how well \\ it represents important information from the dialog\end{tabular} & \begin{tabular}[c]{@{}l@{}}4.03\\ $(\pm0.77)$\end{tabular} \\ \hline
\multirow{2}{*}{\textit{abstractive}} & \begin{tabular}[c]{@{}l@{}}Provide your rating as to the overall \textbf{coverage}\\ of the summary, based on how well\\ it represents important information from the dialog\end{tabular} & \begin{tabular}[c]{@{}l@{}}3.96\\ $(\pm0.84)$\end{tabular} \\ \cline{2-3} 
 & \begin{tabular}[c]{@{}l@{}}Provide your rating as to the \textbf{readability} of \\ the summary. Please consider fluency, grammatical \\ correctness, and coherence\end{tabular} & \begin{tabular}[c]{@{}l@{}}4.22\\ $(\pm0.61)$\end{tabular}
\end{tabular}

}
\caption{Results of the Quality Assessment}
\label{table:datasetQuality}
\end{table}

\subsection{Dataset Analysis}\label{sec:datasetStats}


\begin{table}[]
\centering
\resizebox{0.42\textwidth}{!}{%
\begin{tabular}{llll}
\hline
  & \textit{\textbf{Full dialog}} & \textit{\textbf{Customer utterances}} & \textit{\textbf{Agent utterances}} \\
\hline
\textbf{\#utterances}  & $10.17 (\pm2.31)$  & $5.48 (\pm1.84)$       & $4.69 (\pm1.39)$   \\
\textbf{\#sentences}  & $22 (\pm6.56)$  & $10.23 (\pm4.83)$       & $11.75 (\pm4.44)$   \\
\textbf{\#tokens}   &  $245.01 (\pm79.16)$  & $125.61 (\pm63.94)$       & $119.40 (\pm46.73)$   \\

\end{tabular}
}
\caption{Average lengths of dialogs}
\label{table:length1}
\end{table}

Table \ref{table:length1} details the average length of the dialogs in \datasetName{}, including the average lengths of the customer and agent utterences.  
The average length of the summaries is reported in Table~\ref{table:length2}. 
Comparing the dialog lengths to the summaries lengths indicates the average compression rate of the summaries. For instance, on average, the abstractive summaries compression rate is $85\%$ (i.e. the number of tokens is reduced by $85\%$), while the extractive summaries compression rate is $70\%$. The number of customer and agent sentences selected in the extractive summaries were relatively equally distributed with 7445 customer sentences and 7844 agent sentences in total.

\begin{table}[h]
\centering
\resizebox{0.42\textwidth}{!}{%
\begin{tabular}{llll}
\hline
  & \textit{\textbf{Overall}} & \textit{\textbf{Customer}} & \textit{\textbf{Agent}} \\
\hline
\textbf{Abstractive}  & $36.41 (\pm12.97)$  & $16.89 (\pm7.23)$       & $19.52 (\pm8.27)$   \\
\textbf{Extractive}   &  $73.57 (\pm28.80)$  & $35.59 (\pm21.3)$       & $35.80 (\pm18.67)$   \\

\end{tabular}
}
\caption{Average lengths (in \# tokens) of summaries}
\label{table:length2}
\end{table}






Next, the positions of the sentences selected for the extractive summaries were analyzed. In $85\%$ of the cases, sentences from the first customer utterance were selected, compared to $52\%$ of the cases in which sentences from the first agent utterances were selected. This corroborates the intuition that customers immediately express their need in a typical customer service scenario, while agents do not immediately provide the needed answer: agents typically greet the customer, express empathy, and ask clarification questions.
For the abstractive summaries, inherently, the utterance from which annotators selected information cannot be directly deduced, but can be approximated. Following~\cite{Nallapati_2017},  
for each abstractive summary, we evaluated the ROUGE distance (using ROUGE-L Recall) between the agent (resp. customer) part of the summary, with each of the actual agent (resp. customer) utterances in the original dialog. 
We then considered the utterance with the maximal score to be the utterance from which the summary is mainly based-on. By averaging over all the dialogs, we obtained that $75\%$ of the customer summary part are based-on the first customer utterance vs. only $12\%$ of the agent's part. 
\section{Next Response Prediction Summarizer}\label{sec:NRP_SUMM}
We introduce a novel, unsupervised extractive summarization method (coined \textit{NRP Summ}) aimed at identifying the sentences that influence the entire dialog the most. 
\newline\textbf{The Next Response Prediction Model} - To identify the influence of each sentence on the entire conversation, we utilize the next response prediction (NRP) task \cite{gunasekara2019dstc7} in dialog systems. The NRP task is defined as follows: given a dialog context, i.e., the list of sentences in the dialog up to a certain point ($C = \{s_1, s_2, ..., s_k\}$), predict the next response sentence $(c_r)$  from a given set of candidates $\{c_1,..., c_r,..., c_n \}$. 
To train the NRP model, we used a binary classifier commonly used for GLUE tasks \cite{wang2018glue}.
We process the dialogs to construct triples of \textit{\textless \small{dialog context ($C$), candidate ($c_i$), label ($1/0$)}\textgreater}~from each dialog context.
For each $C$, we create a set of $k+1$ (k=5 in this study) triples: one triple containing the correct response $(c_r)$ (label=$1$), and $k$ triples containing incorrect responses randomly sampled from the dataset (label=$0$).
The dialog context $C$ and a candidate response $c_i$ are fed together to BERT as a sequence {\small ([CLS] $C$ [SEP] $c_i$ [SEP])}.
The hidden state of the [CLS] token was used as the representation of the pair. Training is done using positive and negative examples with cross-entropy loss. A model trained on the NRP task associates a probability $(p_r)$ for the response $(c_r)$, given the context $C$. We trained two NRP models, (1) a model predicting the next response given the prior sentences \textit{(NRP-FW)}, and (2) a model predicting the prior utterance given subsequent utterances \textit{(NRP-BW)}. 
\newline\textbf{Salient sentence identification}- 
The intuition behind this approach is that the removal of the critical sentences from a dialog context will entail a larger drop in probability in predicting a subsequent and prior responses. We follow the hypothesis that the  critical sentences for the NRP task will also be salient sentences for the summary. The sentence removal occurs in two steps. In the initial step, we feed the entire context to the NRP model and identify the probability of predicting the next (or prior) sentence. In the next step, we remove one sentence at a time from the context, and input the new context to the NRP model and identify the probability of predicting the same next (or prior) utterance. Then, we assign the drop in probability as a score to the removed sentence.

To identify the salient sentences in predicting the next response, we remove one sentence at a time from the dialog context $(C \backslash s_i)$ and use that as the input to a trained \textit{NRP-FW} model and identify the probability $(p_r^{fw})$ for the corresponding response $(c_r)$. Then, we assign the drop in probability $(p_r - p_r^{fw})$ as a score to the removed sentence $s_i$ in the context. We follow the same process to identify the drop in probability in predicting the prior sentence, given the same dialog context and masked sentence (using \textit{NRP-BW} model), and assign that as another score for the masked sentence. The averaged score for each sentence is used during salient sentence identification. For the evaluation, we use the top two customer sentences and the two top agent sentences as the extractive summary of the dialog.

\section{Experiments and Results}\label{sec:experiments}
We aim to confirm that \datasetName{} is suitable as a ground-truth dataset for the dialog summarization task.
To this end, we apply and analyze several baseline summarization models as well as \textit{NRP Summ}, to the dataset, as detailed below. We randomly split the dialogs and their associated summaries into three sets: $80\%$ for the training set, $10\%$ for the validation and the rest $10\%$, for the test set.


\subsection{Baselines}\label{sec:baselines}
The baselines evaluated as part of this study are:
\newline \textbf{\textit{Random (extractive)}} - Two random sentences from the agent utterances and two from the customer utterances.
\newline \textbf{\textit{LEAD-4 (extractive)}} - 
The first two sentences from the agent utterances and the first two from the customer utterances.
This approach is considered a very competitive baseline 
(see ~\cite{kryscinski-etal-2019-neural} when considering news summarization).
\newline \textbf{\textit{LexRank (extractive)}} - This unsupervised summarizer \cite{LexRankRadev} casts the summarization problem into a fully connected graph, in which nodes represent sentences and edges represent similarity between two sentences. Pair-wise similarity is measured over the bag-of-words representation of the two sentences. Then, \textit{PowerMethod} is applied on the graph, yielding a centrality score for each sentence. 
We take the two top central customer and agent sentences (2$+$2). 
\newline \textbf{\textit{Cross Entropy Summarizer (extractive)}}- \textit{CES} is an unsupervised, extractive summarizer \cite{RoitmanFCBK20,FeigenblatRBK17}, which considers the summarization problem as a multi-criteria optimization over the sentences space, where several summary quality objectives are considered. The aim is to select a subset of sentences optimizing these quality objectives.
The selection runs in an iterative fashion: in each iteration, a subset of sentences is sampled over a learned distribution and evaluated against quality objectives. 
We introduced some minor tuning to the original algorithm, to suit dialog summarization. First, query quality objectives were removed since we focus on generic summarization.
Then, since dialog sentences tend to be relatively short, when measuring the coverage objective, each sentence was expanded with the two most similar sentences, using Bhattacharyya similarity. Finally, Lex-Rank centrality scores were used as an additional quality objective, by averaging the centrality scores of sentences in a sample. 
\newline \textbf{\textit{PreSumm} (extractive/abstractive)} - This model \cite{liu2019text} 
 applies BERT \cite{devlin2019bert} for text summarization in both extractive and abstractive settings. In the extractive setting, \textit{PreSumm} treats the summarization task as a sentence classification problem: a neural encoder creates sentence representations and a classifier predicts which sentences should be selected for the summary. We used a pre-trained model\footnote{\scriptsize \url{ https://github.com/nlpyang/PreSumm}} and fine-tuned the model using the \datasetName{}. In the abstractive setting, the model uses the same encoder as the extractive model while the decoder is a 6-layered Transformer initialized randomly. 
\newline \textbf{\textit{BART (abstractive)}} - 
A denoising autoencoder \cite{lewis2019bart} that uses the seq2seq transformer architecture. 
It consists of two parts: an encoder
and a decoder. The encoder is a bidirectional encoder
which corresponds to the structure of BERT, and the decoder is an auto-regressive decoder following the settings of GPT \cite{radford2019language}. We use a lightweight variant of \textit{BART} (coined \textit{DistilBART}) 
that is fine-tuned on the XSum task \cite{narayan2018don}.
We further fine-tuned the model using the \datasetName{}.
Different variants of the BART model that were evaluated are discussed in the results section. The hyper-parameters are described in the supplemental material.


\begin{table}[t]
\caption{ROUGE F-Measure evaluation on the test set, supervised baselines are marked with $\dagger$}%
\centering
\resizebox{0.42\textwidth}{!}{%
\begin{tabular}{llcccc}
\hline
\textit{\begin{tabular}[c]{@{}l@{}}Length\\ Limit\end{tabular}} & \textit{Method Name} & \textit{R-1} & \textit{R-2} & \textit{R-SU4} & \textit{R-L} \\ \hline
\multicolumn{6}{c}{\textit{Abstractive Dataset}} \\ \hline
\multirow{11}{*}{\begin{tabular}[c]{@{}l@{}}\textit{35}\\ \textit{tokens}\end{tabular}} 
 & \textit{Random} & 22.970 & 6.370 & 8.340 & 20.601 \\
 & \textit{Lead} & 26.666 & 10.098 & 11.690  & 24.360 \\
 & \textit{LexRank} & 27.661 & 10.448 & 12.249 & 24.900 \\
 & \textit{CES} & 29.105 & 11.483 & 13.344 & 26.281 \\
  & \textit{NRP Summ} & \textbf{30.197} & \textbf{12.219} & \textbf{13.911} & \textbf{27.111}  \\
 & \textit{BART - without fine-tuning} & 20.365 & 4.110 & 6.188 & 16.019\\
 \cline{2-6}
 & \textit{PreSumm extractive $\dagger$} & 30.821 & 12.972 & 14.633 & 27.909 \\
 & \textit{PreSumm abstractive $\dagger$} & 33.468 & 9.284 & 13.115 & 31.003\\
 & \textit{BART - without ext $\dagger$} & 36.395  & 18.015 & 18.346  & 32.280  \\
 & \textit{BART - with ext $\dagger$} & \textbf{38.237} & \textbf{19.449} & \textbf{19.594} & \textbf{33.818}  \\ \hline
\multirow{11}{*}{\begin{tabular}[c]{@{}l@{}}\textit{70}\\ \textit{tokens}\end{tabular}} & \textit{Random} & 26.930 & 8.870 & 10.980 & 24.337 \\
 & \textit{Lead} & 28.913  & 11.489 & 13.053  & 26.395 \\
 & \textit{LexRank} & 30.457 & 12.379 & 14.202 & 27.486 \\
 & \textit{CES} & \textbf{31.465} & 13.152 & \textbf{14.954} & \textbf{28.464} \\
 & \textit{NRP Summ} & 31.416 &\textbf{ 17.365} & 14.043 & 27.623 \\
  & \textit{BART - without fine-tuning} & 20.378 & 4.127 & 6.200 & 16.028\\
  \cline{2-6}
 & \textit{PreSumm extractive $\dagger$} & 33.220  & 14.288 & 15.986 & 30.305\\
 & \textit{PreSumm abstractive $\dagger$} & 33.010 & 9.493 & 12.974 & 30.667   \\
 & \textit{BART - without ext $\dagger$} & 36.076  & 17.844 & 18.161  & 31.939  \\
 & \textit{BART - with ext $\dagger$} & \textbf{37.938} & \textbf{19.263} & \textbf{19.417} & \textbf{33.508}   \\ \hline
\multirow{12}{*}{\begin{tabular}[c]{@{}l@{}}\textit{unlimited} \\ \end{tabular}} 
 & \textit{Random} & 26.865 & 8.848 & 10.946 & 24.269 \\
 & \textit{Lead} & 29.061 & 11.560  & 13.106   & 26.470 \\
 & \textit{LexRank} & 30.459 & 12.652 & 14.423  & 27.563 \\
 & \textit{CES} & \textbf{31.569} & 13.334 & 15.118 & \textbf{28.552} \\
 & \textit{NRP Summ} & 31.209 & \textbf{17.265} & \textbf{17.956} & 28.541 \\
  & \textit{BART - without fine-tuning} & 20.378 & 4.127 & 6.200 & 16.028\\
  \cline{2-6}
 & \textit{PreSumm extractive $\dagger$} & 32.815 & 14.149 & 15.799 & 30.026 \\
 & \textit{PreSumm abstractive $\dagger$} & 33.001 & 9.494 & 12.971 & 30.650 \\ 
 & \textit{BART - without ext $\dagger$} & 36.076  & 17.844 & 18.161  & 31.939  \\
 & \textit{BART - with ext $\dagger$} & \textbf{37.938}  & \textbf{19.263 } & \textbf{19.417}  & \textbf{33.508}  \\ \hline
\multicolumn{6}{c}{\textit{Extractive Dataset}} \\ \hline
\multirow{6}{*}{\begin{tabular}[c]{@{}l@{}} \textit{35} \\ \textit{tokens}\end{tabular}} 
 & \textit{Random} & 32.761 & 17.843 & 17.794 & 30.518 \\
 & \textit{Lead} & 53.156 & 42.944  & 40.549  & 52.045  \\
 & \textit{LexRank} & 48.584 & 36.758 & 36.125 & 46.847 \\
 & \textit{CES} & 55.328  & 45.032 & 43.841 & 54.182 \\
  & \textit{NRP Summ} & \textbf{58.410} & \textbf{49.490} & \textbf{47.404}  & \textbf{57.428}  \\
   \cline{2-6}
 & \textit{PreSumm extractive $\dagger$} & \textbf{60.957} & \textbf{52.478} & \textbf{50.908} & \textbf{60.142} \\
 \hline
\multirow{6}{*}{\begin{tabular}[c]{@{}l@{}}\textit{70}\\ \textit{tokens}\end{tabular}} 
 & \textit{Random} & 47.868 & 32.978 & 32.693 & 46.035 \\
 & \textit{Lead} & 57.491 & 47.199  & 45.388   & 56.531  \\
 & \textit{LexRank} & 55.773 & 43.365 & 42.563  & 54.290 \\
 & \textit{CES} & 58.984 & 47.713 & 46.387 & 57.889 \\
 & \textit{NRP Summ} & \textbf{61.114}  &\textbf{ 51.381} & \textbf{49.558} & \textbf{60.292}  \\
  \cline{2-6}
 & \textit{PreSumm extractive $\dagger$} & \textbf{65.158} & \textbf{55.813} & \textbf{53.517} & \textbf{64.370}  \\
 \hline
 \multirow{6}{*}{\begin{tabular}[c]{@{}l@{}}\textit{unlimited}\\ \end{tabular}} 
 & \textit{Random} & 48.943 & 35.074 & 34.548 & 47.333 \\
 & \textit{Lead} & 54.995 & 44.425  & 42.796    &  53.943  \\
 & \textit{LexRank} & 57.018 & 45.332 & 44.459   &  55.772\\
  & \textit{CES} & 59.872 & 49.126 & 47.722 & 58.874 \\
 & \textit{NRP Summ} & \textbf{62.971} & \textbf{55.411} & \textbf{54.614} & \textbf{62.596}  \\
  \cline{2-6}
 & \textit{PreSumm extractive $\dagger$} & \textbf{65.659} & \textbf{56.628} & \textbf{54.327} & \textbf{64.943}  \\
 \hline
\end{tabular}}
\label{tab:ROUGE-F}
\end{table}

\subsection{Automatic Evaluation}
We first use automatic measures to evaluate the summaries generated by the 
models described above, using 
the reference summaries of \datasetName{}. 
We measured summarization quality using the ROUGE measure~\cite{lin2004rouge} compared to the ground truth. We use the official toolkit with its standard parameters setting\footnote{\tiny ROUGE-1.5.5.pl -a -c 95 -m -n 2 -2 4 -u -p 0.5}. For the limited length variants, we run ROUGE with its limited length constraint. 
Table~\ref{tab:ROUGE-F} reports ROUGE F-Measure results. We evaluate all summarization models (extractive and abstractive, where the extractive summarizers are set to extract 4 sentences) against the abstractive and extractive datasets. Supervised baselines are marked with the $\dagger$ symbol. Based on the average length of the summaries, reported in Table~\ref{table:length2}, we evaluate ROUGE with three length limits: $35$ tokens (the average length of the abstractive summaries), $70$ tokens (the average length of the extractive summaries) and \textit{unlimited}. Below we discuss these results in detail.


\subsubsection{\datasetName{} Abstractive Dataset} \label{sec:abs_results}
\textbf{Quality of extractive summarization models}- We start by analyzing how well extractive summarization models perform on the abstractive 
reference summaries. As described in Table~\ref{tab:ROUGE-F}, we note that in most cases, except 70 tokens summary, \textit{NRP Summ} outperforms other unsupervised, extractive baselines. 
Interestingly, the performance of the simple \textit{Lead-4} baseline is not far from that of the more complex unsupervised baselines. 
For instance, considering the 70 tokens results of the abstractive dataset, \textit{LexRank} outperforms \textit{Lead-4} by only $4\%$-$8\%$.  This is backed up by the statistics we report in section~\ref{sec:datasetStats}, namely that salient content conveyed by the customer appears at the beginning of the dialog. To rule out any potential overfitting, we also present results of the unsupervised, extractive, summarizers against the validation set. Table~\ref{tab:ROUGE-F-validation} shows a similar trend: in most cases, \textit{NRP Summ} outperforms  other models.

\textbf{Quality of abstractive summarization models}- We analyze three variants of the BART model: \textbf{(1)} \textit{BART} with no fine-tuning on \datasetName{} (\textit{BART-without-fine-tuning}), \textbf{(2)} BART fine-tuned on \datasetName{} (\textit{BART-without-ext}), \textbf{(3)} BART fine-tuned on \datasetName{} with the extractive summary provided as input in addition to the dialog (\textit{BART-with-ext}). For training the \textit{BART-with-ext}, the ground truth extractive summaries were appended to the dialog (with a dedicated separator). For validation and testing, the extractive summaries generated by the \textit{NRP Summ} model were used. All BART models were pre-trained on the XSum summarization dataset~\cite{xsum-emnlp} (see the specific system models settings in the supplemental material). As described in Table~\ref{tab:ROUGE-F}, the BART models fine-tuned on \datasetName{} obtain the best results by far, compared to all other models.  \textit{BART-without-fine-tuning} model performs poorly, compared to all the other models. 
\textbf{From this analysis we learn that, pre-training on the general summarization task is not sufficient, fine-tuning is required to help the  model learn the specifics of the  dialog summarization task}. Interestingly, \textit{BART-with-ext} outperforms \textit{BART-without-ext}, suggesting that the extractive summary helps the model to attend to salient content.
Although the PreSumm model was also similarly 
fine-tuned on \datasetName{}, its performance is inferior to BART.

\subsubsection{\datasetName{} Extractive Dataset}  Here we focus on evaluating the extractive summarizaion models on the extractive dataset. We first note that the average length of ground truth extractive summaries in \datasetName{} is 4 sentences out of 22 sentences, on average, in a dialog. 
The lower compression rate of the extractive summaries compared to the abstractive summaries leads to higher ROUGE scores of the extractive summaries. The \textit{NRP Summ} model outperforms all unsupervised methods, while the supervised \textit{PreSumm extractive} model outperforms all other models.

\subsection{Human Evaluation}
We conducted two human evaluation studies to assess the quality of the summarization models. 
The first focuses on the \textbf{Informativeness} and \textbf{Saliency} of the summaries generated by the models. Following~\cite{DBLP:conf/emnlp/LiuL19, DBLP:conf/acl/LiuL19}, we used the QA paradigm to test whether the summarization models retain key information. We chose to evaluate the two abstractive models \textit{BART-without ext} and \textit{PreSumm-abs} and four extractive models - \textit{NRP Summ}, \textit{CES}, \textit{PreSumm-ext} and \textit{LEAD} (limited to 4 sentences). We randomly selected 20 dialogs and recruited 4 NLP expert annotators for the task. One was asked to create a set of questions based on the three ground truth abstractive summaries from \datasetName{}, and the other three were asked to read the generated summaries and answer the questions.
Using the abstractive rather than the extractive summaries allows the questions to focus on the most salient information, since the extractive summaries are constrained by having a limit of sentences selected as-is from the dialog. 
For each dialog, 4$-$10 yes/no questions regarding the information included in the summary (e.g. \textit{``Does the summary specify that ...''}), were created by the human annotator. Following~\cite{DBLP:conf/naacl/NenkovaP04}, we assigned each question a weight, $w_j$ which is the ratio of ground-truth summaries containing an answer for question $j$. Clearly, important information should be included in several human summaries. 
Then, the other three annotators, $i\in\{1,2,3\}$ were given the set of questions and one summary at a time (without knowing which model generated the summary)
, and were asked to indicate whether the summary contained an answer to the question. Denote the indicator $I_{ij}$ to be $1$ if annotator $i$ determined that the summary contained an answer to question $j$, and $0$ otherwise. The score of a summary generated by a model per dialog $d$ is calculated as $S_d=(100/(3*\sum_{j=1}^{K_d} w_j))\sum_{i=1}^3\sum_{j=1}^{K_d} w_j*I_{ij}$, where $K_d$ is the number of questions given $d$. The highest score a summary can get is 100 which occurs when all annotators agreed that the summary includes the information in all questions. Refer to the supplemental material for examples of questions that were created as part of this evaluation. 

Table~\ref{tab:humanEvalQA} reports the evaluation results, when calculating the summary scores separately for questions pertaining to the agent and customer utterances. 
The Lead-4 baseline outperforms other methods for summarizing customer utterances, which is expected as remarked in sub-section~\ref{sec:abs_results}. In this case, the simple baseline is hard to beat. However, for summarizing agent  utterances, the more advanced models are better, but even the supervised \textit{PreSumm} and \textit{BART} models leave much room for improvement.

 
Following~\cite{DBLP:conf/acl/LiuL19}, we further assess the quality of the summaries along the two dimensions of \textit{Readability} and \textit{Informativeness}. We chose to evaluate only the abstractive models (\textit{BART-without ext} and \textit{PreSumm}) since a high level of \textit{Readability} is not expected with extractive summaries. The annotators were asked to indicate which summary is better with respect to their \textit{Readability} and \textit{Informativeness}, without knowing which system was used to generate which summary. In more than $90\%$ of the cases \textit{BART} outperforms \textit{PreSumm} on both dimensions, consistent with the results in Table~\ref{tab:humanEvalQA}. 


\subsection{Further Analysis of BART summaries}\label{ft-bart}
In section \ref{sec:abs_results} we showed that fine tuning BART on \datasetName{} significantly improves the summaries compared to using BART with no fine tuning. Here we examine, whether using \datasetName{} for fine tuning improves BART's ability to learn an important characteristic of dialog summarization, namely, that a summary should convey text from both speakers (agent and customer). 
We consider three variants of BART: \textbf{(1)} BART  fine tuned on \datasetName{}, \textbf{(2)} BART fine tuned as in (1) for which additional speaker tags (agent or customer) were added  during fine tuning, \textbf{(3)} original BART variant, with no fine-tuning on \datasetName{}. 
We generate summaries for each dialog in the test set using each of the aforementioned variants (1)-(3). Following~\cite{Nallapati_2017},  for each generated summary, we find the two dialog utterances which are most similar to it, using ROUGE-L Recall, and ask whether they represent both speakers, or only one of them. 
We find that in $78\%$ and $79\%$ of cases, both speakers are represented for variants \textbf{(1)} and \textbf{(2)} respectively, but in only $46\%$ of the cases for variant \textbf{(3)}. These should be compared to the baseline of choosing two random utterances, where in $58\%$ of the cases both speakers are represented.
The differences of distribution between variants \textbf{(1)} and \textbf{(2)}, compared to variants \textbf{(3)} (as well as the random baseline) are statistically significant~{\small (Welch Two Sample t-test, $p$<$10^{-6}$)}. \textbf{This analysis strengthens the confidence we have in \datasetName{} and the ability to use it for the dialog summarization tasks}. 

\begin{table}[t]
\caption{ROUGE F-Measure on validation set}%
\centering
\resizebox{0.37\textwidth}{!}{%
\begin{tabular}{llcccc}
\hline
\textit{\begin{tabular}[c]{@{}l@{}}Length\\ Limit\end{tabular}} & \textit{Method Name} & \textit{R-1} & \textit{R-2} & \textit{R-SU4} & \textit{R-L} \\ \hline
\multicolumn{6}{c}{\textit{Abstractive Dataset}} \\ \hline
\multirow{5}{*}{\begin{tabular}[c]{@{}l@{}}35\\ tokens\end{tabular}} & \textit{Random} & 24.459 & 7.719 & 9.504 & 22.157 \\
 & \textit{Lead} & 28.569  & 11.623 & 13.058 & 26.088 \\
 & \textit{LexRank} & 27.039  & 10.120  & 12.030 & 23.990 \\
  & \textit{CES} & 30.693  & 13.129  & 14.752 & 27.606 \\
 & \textit{NRP} & \textbf{30.889}  & \textbf{13.410} & \textbf{14.901} & \textbf{27.890} \\
  \hline
\multirow{5}{*}{\begin{tabular}[c]{@{}l@{}}70\\ tokens\end{tabular}} & \textit{Random} & 28.249 & 10.480 & 12.277 & 25.721 \\
 & \textit{Lead} &  31.127 &  13.536 & 14.867   & 28.542  \\
 & \textit{LexRank} & 30.302 & 12.444 & 14.161 & 27.191 \\
  & \textit{CES}& \textbf{32.769} & 14.125 & \textbf{15.650} & \textbf{29.516}     \\
 & \textit{NRP} & 32.453 & \textbf{14.694} & 15.316 & 29.119 \\
 \hline
\end{tabular}}
\label{tab:ROUGE-F-validation}
\end{table}

\begin{table}[t]
 \caption{System scores based on questions answered}\label{tab:humanEvalQA}\centering
\resizebox{0.36\textwidth}{!}{%
\begin{tabular}{llcc}
\textbf{Model}   & \textbf{Type} & \textbf{Customer} & \textbf{Agent} \\ \hline
LEAD             & ext.           & 77.9              & 39.2           \\
CES              & ext.           & 69.6              & 49.9           \\
NRP Summ         & ext.           & 71.3              & 40.8           \\
PreSumm$\dagger$          & ext.           & 74.3              & 51.2           \\ \hline
PreSumm$\dagger$          & abs.           & 16.0              & 12.5           \\
BART-without-ext$\dagger$ & abs.           & 58.5              & 31.7          
\end{tabular}
}
\end{table}


\section{Related Work}\label{sec:related}
\textbf{Document Summarization}-
Text summarization has been studied for many years and several public datasets have been published in this domain. 
One central problem in summarization research is 
the high cost of generating ground truth data.
Whereas, in some datasets, such as DUC ~\cite{dang2005overview} and Xsum~\cite{xsum-emnlp}, reference summaries were created specifically for the dataset, 
in other works different strategies are employed to identify existing texts that can be used as reference summaries.
For example, in the case of single-document summarization, the CNN/Dailymail 
the key points associated with published news articles as part of the editorial process~\cite{DBLP:journals/corr/NallapatiXZ16}, are taken to be the reference summary of the news article.
Other datasets, such as NewsRoom, Gigaword, NYT,  ~\cite{grusky2018newsroom,rush2015neural, sandhaus2008new} also focus on the news domain, leveraging existing texts as reference summaries.
Summarization of scientific articles has also been studied as in~\cite{yasunaga2019scisummnet}, treating abstracts as well as sentences describing another paper, as potential reference summaries.


\textbf{Data Driven Dialog Systems}-
Many aspects of data driven dialog systems have undergone a revolution in recent years with the advent of ever more powerful techniques based on deep learning~\cite{serban2016building, henderson2019training, zhang2019dialogpt, wu2020controllable}.
Most of the available dialog datasets support dialog tasks such as next response prediction \cite{kadlec2015improved, bordes2016learning, byrne2019taskmaster}, conversational question answering \cite{reddy2019coqa, choi2018quac, saeidi2018interpretation} and dialog state tracking \cite{budzianowski2018multiwoz, rastogi2019towards}.


\textbf{Dialog Summarization Datasets}-
On the other hand, summarization of two-party dialogs is relatively unexplored due to the lack of suitable large scale benchmark data. Most of the previous works on abstractive dialog summarization~\cite{banerjee2015abstractive, mehdad2014abstractive, goo2018abstractive, li2019keep}  focus on the AMI meeting corpus dataset~\cite{McCowan2005AMI}. 
This dataset has multiple deficiencies including, its size (only $141$ summaries are available), and the quality of the ground truth summaries, since the meeting description is treated as the summary.
The Argumentative Dialog Summary Corpus \cite{misra2015using}, a small dataset of $45$ dialogs, is based on political debates from the Internet Argument Corpus \cite{walker2012corpus} where summaries are constructed by crowd-workers. 
More recently, CRD3 \cite{rameshkumar2020storytelling} was introduced, a spoken conversation dataset that consists of $159$ conversations and summaries. The SAMSum dialog corpus \cite{gliwa2019samsum} contains over $16$k chat conversations with manually annotated abstractive summaries. 
However, this dataset contains role-playing open domain, {\it chichat} dialogs, and does not provide ground truth for extractive summarization. 
In contrast, \datasetName{} involves different summarization challenges, e.g, identifying problems and provided solutions. 
\cite{yuan2019abstractive} studied the problem of abstractive dialog summarization using a dataset constructed from the  MultiWOZ-2.0 dataset \cite{budzianowski2018multiwoz}. 
This dataset considers the instructions provided to crowd-workers as part of the Wizard-of-OZ setting as the ground truth summary. 
Hence, the dataset does not contain ``real'' summary annotations for dialogs. \cite{liu2019automatic} worked on the problem of automatic summary generation for customer service dialogs, but the dataset is not publicly available. 
Recently, MediaSum~\cite{zhu2021mediasum} was released, suggesting the use of overview and topic descriptions as summaries of $460$k interview transcripts from NPR radio channel. 

\section{Conclusion}\label{sec:conclusion}

In this paper, we release \datasetName{}, the first open large-scale dataset focused on summarization of customer-support dialogs.
We conducted automatic and human evaluation studies to ensure the high-quality of the human-generated extractive and abstractive summaries. 
To test the applicability of the dataset, we evaluated 
various baselines, as well as a new extractive summarization method, \textit{NRP Summ}, and showed that while automatically generated abstractive summaries achieve high quality, there is still much room for improvement. 
We believe \datasetName{} will help foster research in this real-world scenario, which was previously little studied due to lack of suitable datasets.

\renewcommand{\UrlFont}{\ttfamily\small}
\section{Ethics}
We constructed \datasetName{} dialogs using the publicly available \textit{Customer Support on Twitter} dataset (\url{ www.kaggle.com/thoughtvector/customer-support-on-twitter}). The summaries generation task was executed on Appen.com platform; we only recruited 
crowd-workers that are members of an Expert Business Partner channel, 
fluent English speakers, with a very high 
approved task acceptance rate. 
We have set the task payment, so that crowd-workers are expected to earn 9\$ per hour. 

\bibliography{anthology,bibfile}
\bibliographystyle{acl_natbib}

\appendix

\section{\datasetName{} Dataset - Summaries Generation}
\begin{figure*}[t!]
     \centering
     \includegraphics[width=0.6\textwidth,angle=-90]{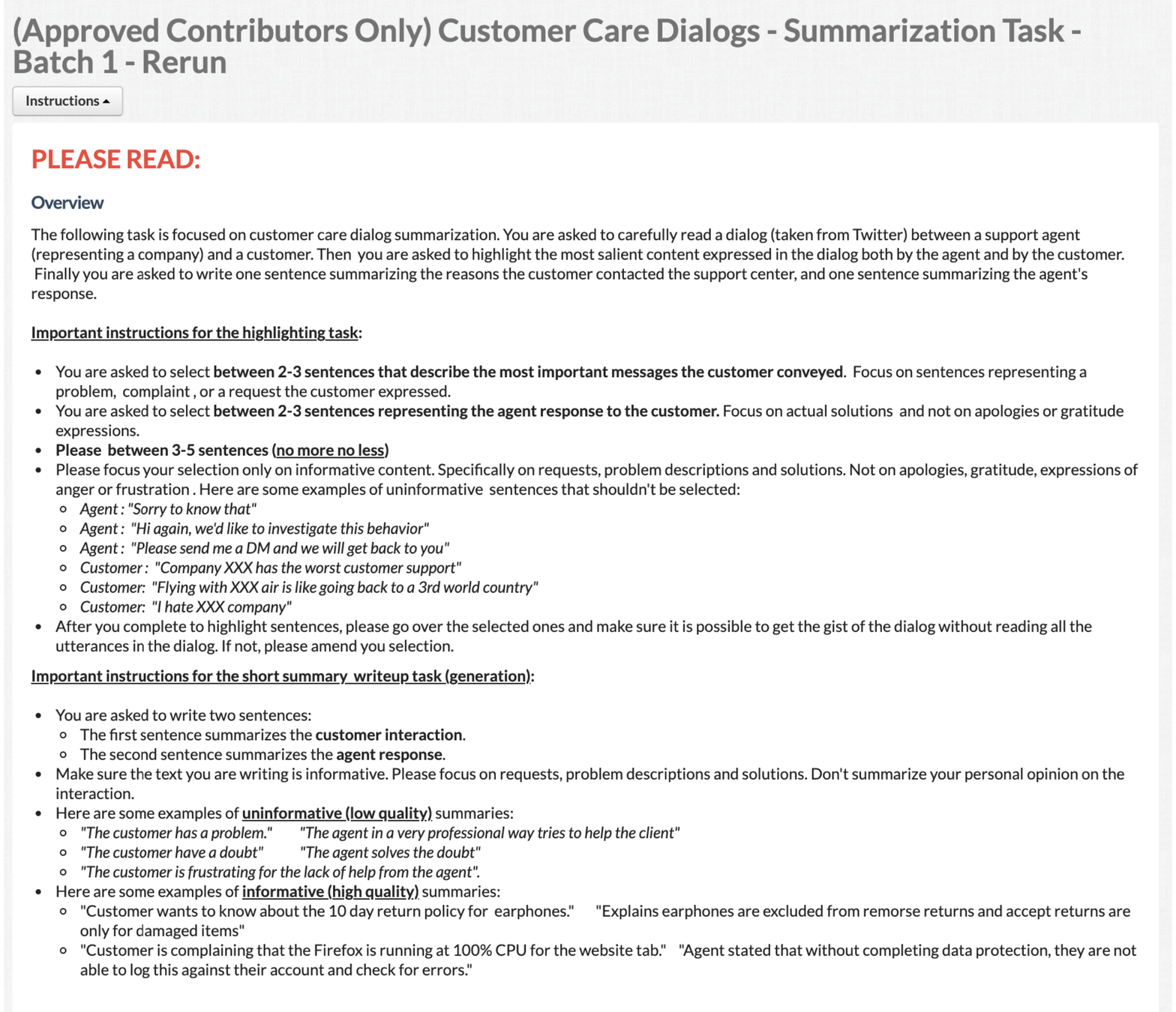}
     \caption{Annotation interface - Instructions for the summary generation task}
     \label{fig:appen_instruct}
 \end{figure*}

\begin{figure*}[t!]
     \centering
     \includegraphics[width=0.7\textwidth,angle=-90]{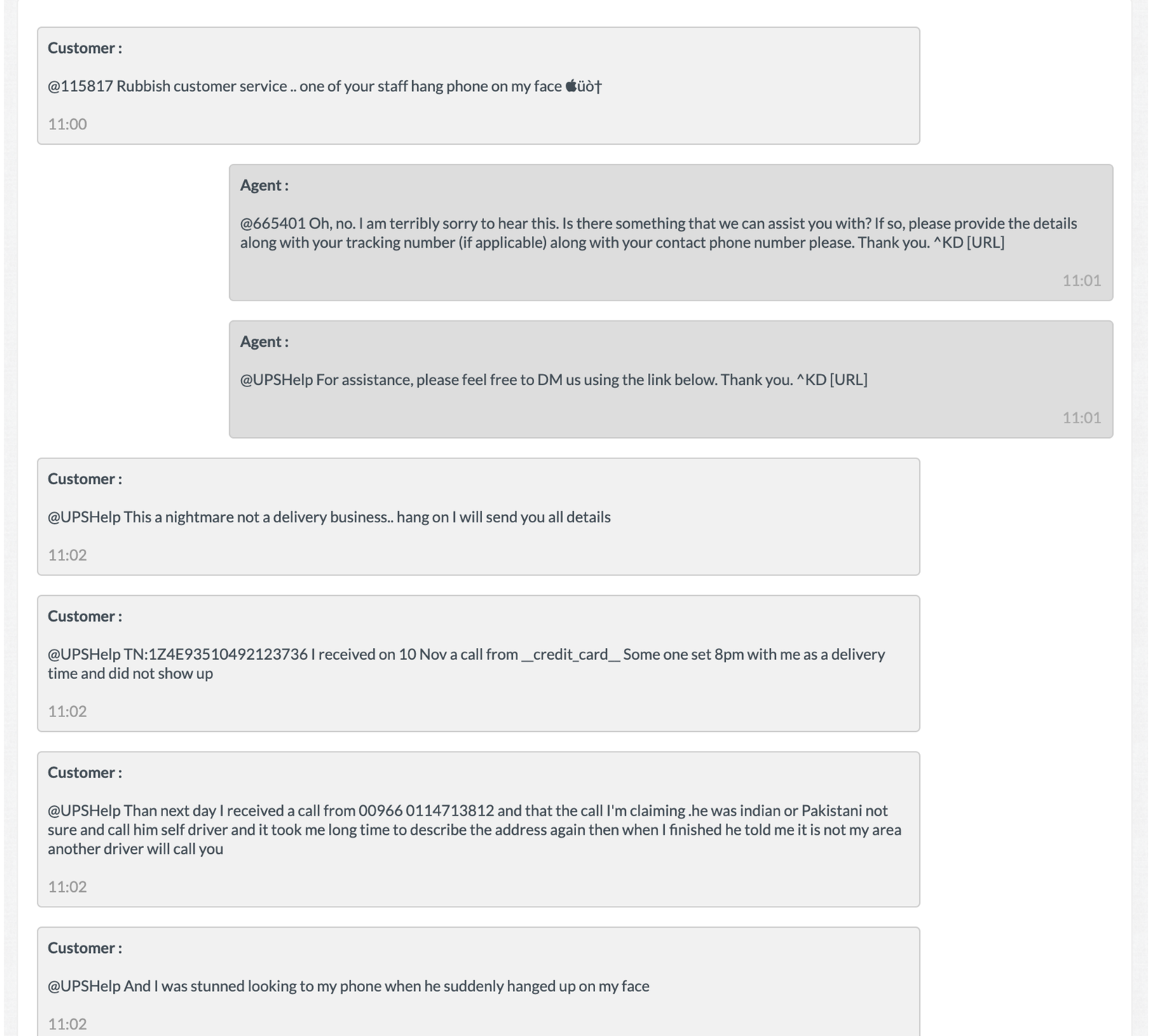}
     \caption{Annotation interface - A dialog presented to annotators}
     \label{fig:appen_dialog}
 \end{figure*}

\begin{figure*}[t!]
     \centering
     \includegraphics[width=0.6\textwidth,angle=-90]{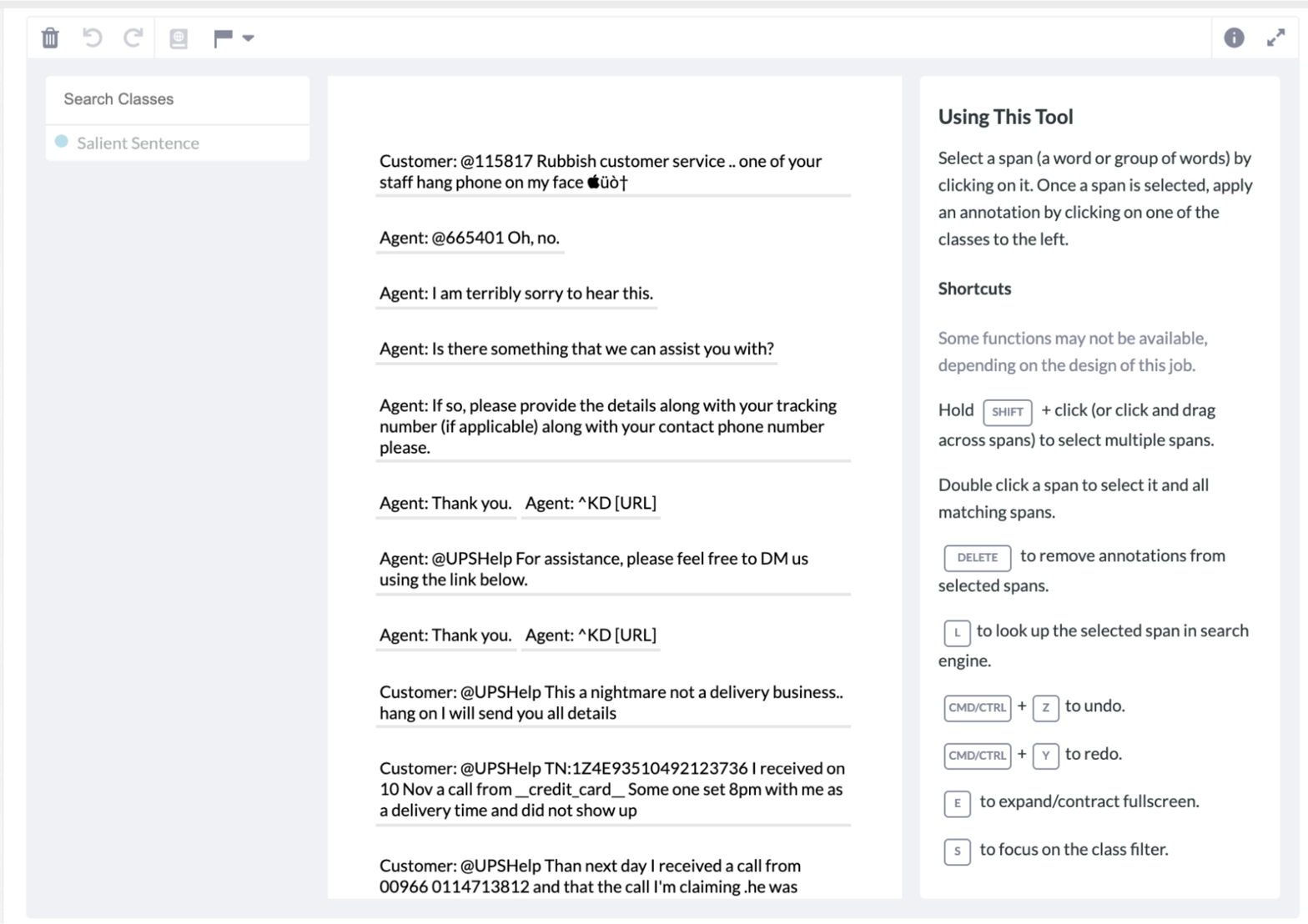}
     \caption{Annotation interface - Annotators are asked to highlight salient sentences (for the extractive summary)}
     \label{fig:summary_generation}
 \end{figure*}

As described in the main paper, \datasetName{} dialogs were sent for generation of summaries using crowd-sourcing on the Appen.com platform. Figure \ref{fig:appen_instruct} shows the instructions provided to annotators working on \datasetName{} summary generation task. Figure \ref{fig:appen_dialog} shows how dialogs were presented to annotators as part of the annotation interface. Figure \ref{fig:summary_generation} shows the dialog annotation interface: annotators are asked to highlight the salient sentences (extractive summary) in the dialog. 
In the following sub-sections we describe in details the instructions crowd-workers received while working on this task. 

\subsection{Extractive Summaries} 
The annotators were asked to select $2$ to $3$ entire sentences that describe the most important messages the customer conveyed. They were asked to focus on sentences presenting a problem, complaint, or a request the customer expressed. Then, they were asked to select between $2$ to $3$ entire sentences representing the agent response to the customer, with focus on actual solutions and not on apologies or gratitude expressions. Clearly, the analysis of the emotional part of customer interactions is also important. However, this is associated with other NLP tasks such as sentiment analysis. The same decision was taken in~\cite{liu2019automatic}. As a final step, the annotators were asked to go over the selected summary sentences and make sure that they represent the full dialog as much as possible. 
In addition, several examples of uninformative sentences, that should not appear in summaries, were given to help annotators understand the requirements better (e.g. {\it ``We're sorry to hear that.'', ``Poor customer service.'', ``Hi again, we'd like to investigate this behavior.'', ``I hate X company''}).

\subsection{Abstractive Summaries}
Here, the annotators were instructed to write two sentences summarizing the whole dialog, one summarizing the customer questions/requests and the second one summarizing the agent responses. We limited ourselves to two sentences to simplify the task of the crowd-workers. In addition, having separate summary sentences  allow an automated summarizer to (potentially) generate two summaries, one for the customer and one for the agent.  Similarly to the extractive summarization, annotators were asked to write an informative summary, that focuses on requests, problem descriptions and solutions excluding personal opinions, insults or apologies.

\section{Model Training and Hyperparameter Details}

In this section, we elaborate the training processes and the hyperparameters used in the supervised trained models used in this study. Each experiment was run on 2 V100 GPUs (on
a single machine).

\subsection{Next response prediction model for NRP Sum}
As introduced in the main paper, the NRP Sum model uses a BERT based binary classifier. The code will be open-sourced in a public git page upon paper acceptance. For this task, we used the \textit{BertForSequenceClassification} model of HuggingFace \cite{wolf2019huggingface}, commonly used for GLUE tasks 
\cite{wang2018glue}. We process the dataset to construct triples of \textless dialog context ($C$), candidate ($c_i$), label ($1/0$)\textgreater~from each dialog context.
For each $C$, we create a set of $10$ triples: one triple containing the correct response (label=$1$), and $9$ triples containing incorrect responses randomly sampled from the dataset (label=$0$). Training is done using positive and negative examples with cross-entropy loss. 

The hyperparameters used for training the model are as follows:

\begin{Verbatim}[fontsize=\small]
model=bert-base-cased
do_lower_case=True
max_seq_length=512
per_gpu_eval_batch_size=24
per_gpu_train_batch_size=24
learning_rate=2e-5
num_train_epochs=5
adam_epsilon=1e-8 
max_grad_norm=1.0 
\end{Verbatim}

We trained two models with this approach, one for predicting the next response given a dialog context and, another to predict the previous sentence given the dialog context. The results of the two models on the validation set are shown in Table 1. 

\begin{table}[ht]
\begin{center}
\begin{tabular}{lccccc}
    \hline
    \multicolumn{1}{p{1cm}}{\centering Model} & 
    R@1 & 
    R@2 & 
    R@5 
\\ \hline
    NRP & 56.09 & 75.95 & 98.08 \\ \hline
    PRP & 51.91 & 73.51 & 95.64 \\ \hline

\end{tabular}
\label{NRP_res}
\end{center}
\caption{The results of the next response prediction task. The model NRP refers to the task of predicting the next response given a dialog context, and the model PRP refers to the task of predicting the previous response given a dialog context.}
\end{table}

\subsection{PreSumm model}
The PreSumm \cite{liu2019text} model was used as a baseline in this study. We used the PreSumm extractive summarization model which was pre-trained on the CNN/DM summarization dataset, and fine-tuned the model on the \datasetName dataset. All the code and pre-trained models used in this study are publicly available\footnote{\url{https://github.com/nlpyang/PreSumm}}. 

The hyperparameters used for training the extractive summarization model are as follows:

\begin{Verbatim}[fontsize=\small]
ext_dropout=0.1
lr=2e-3
save_checkpoint_steps=5000
batch_size=3000
train_steps=50000
accum_count=2
warmup_steps=10000
max_pos=512
\end{Verbatim}

The checkpoint which produced the best performance on the validation dataset (checkpoint at step 35000) was used to initialize the PreSumm abstractive summarization model.  The hyperparameters used for training the abstractive summarization model are as follows:

\begin{Verbatim}[fontsize=\small]
dec_dropout=0.2
sep_optim=true
lr_bert=0.002
lr_dec=0.2
save_checkpoint_steps=5000
batch_size=140
train_steps=100000
accum_count=5
use_bert_emb=true
use_interval=true
warmup_steps_bert=20000
warmup_steps_dec=10000
max_pos=512
beam_size=5
\end{Verbatim}

The checkpoint which produced the best performance on the validation dataset (checkpoint at step 55000) was used to generate summaries on the test dataset. 

\begingroup
\renewcommand{\arraystretch}{0.6} 
\begin{figure}[ht]
\centering
\resizebox{0.47\textwidth}{!}{%
\begin{tabular}{p{0.7cm}p{6.7cm}}
\hline
\\
\multicolumn{2}{c}{{\textit{An awful smell in a flight}}} \\
\hline
\hline
\multicolumn{2}{l}{\tiny{\textbf{Ground truth (human) abstractive summary}}} \\
\hline
& \tiny{Customer complains about smell in flight. Agent updated the customer to seek further assistance by reaching out to one of their in-flight crew members on duty.}\\
\hline
\hline
\multicolumn{2}{l}{\tiny{\textbf{Sample QA Questions}}} \\
\hline

&\tiny{Does the summary specify the customer is complaining about bad smell in his flight?}\\
&\tiny{Does the summary specify the agent asked to contact in-flight crew member on duty for assistance?}\\
&\tiny{Does the summary specify the customer asked to change seat in rebooking?}\\
&\tiny{Does the summary specify the agent apologized for the discomfort?}\\
\hline
\hline
\multicolumn{2}{l}{\tiny{\textbf{Automated abstractive summary}}} \\
\hline
\tiny{\textit{BART}}&\tiny{Customer is complaining about the smell on flight 1287 from Miami to LaGuardia. Agent requests to reach out to a flight attendant to address the odor in the aircraft.}\\
\hline
\hline
\multicolumn{2}{l}{\tiny{\textbf{Automated extractive summaries}}} \\
\hline
\tiny{\textit{NRP}} & \tiny{\textbf{Customer}} \tiny{Flight1287 from Miami to LaGuardia smells awful. Every person getting on the flight is complaining.}\\
  & \tiny{\textbf{Agent}} {Unfortunately, our First Class Cabin is full on our DL1287 flight for today, Allie. Please reach out to a flight attendant to address the odor in the aircraft.}\\
\hline
\tiny{\textit{LEAD}} & \tiny{\textbf{Customer}} \tiny{Flight1287 from Miami to LaGuardia smells awful. It’s really really bad.}\\
  & \tiny{\textbf{Agent}} \tiny{Allie, I am very sorry about this. Please reach out to a flight attendant to address the odor in the aircraft.}\\
\hline
\tiny{\textit{CES}} & \tiny{\textbf{Customer}} \tiny{Flight1287 from Miami to LaGuardia smells awful. They told us to rebook, then told us the original flight was still departing.}\\
  & \tiny{\textbf{Agent}} \tiny{Unfortunately, our First Class Cabin is full on our DL1287 flight for today, Allie. You may seek further assistance by reaching out to one of our in-flight crew members on duty.}\\
\\
\hline
\\

\multicolumn{2}{c}{{\textit{A Red Eye Removal issue}}} \\
\hline
\hline
\multicolumn{2}{l}{\tiny{\textbf{Ground truth (human) abstractive summary}}} \\
\hline
& \tiny{Customer is asking help how to remove red eye in Ligthroom CC since he can't find it in tool, and customer wants some new advanced features. Agent is giving details on it, then sends a link where he can get help and also asks customer to report a complaint and his engineer team will get alert and help him over it.}\\
\hline
\hline
\multicolumn{2}{l}{\tiny{\textbf{ Sample QA Questions}}} \\
\hline

&\tiny{Does the summary specify the customer asks to do red eye removal?}\\
&\tiny{Does the summary specify the customer is using Lightroom CC?}\\
&\tiny{Does the summary specify the agent sent an article containing the required information?}\\
&\tiny{Does the summary specify the agent explained the released version contains all the features of the old version?}\\

&\tiny{Does the summary specify the agent suggested the customer to report a complaint so the engineering team will get an alert and help?}\\
\hline
\hline
\multicolumn{2}{l}{\tiny{\textbf{Automated abstractive summary}}} \\
\hline
\tiny{\textit{BART}}&\tiny{Customer is asking how to do red eye removal in Lightroom CC. Agent is looping their expert team to help answer the question.}\\
\hline
\hline
\multicolumn{2}{l}{\tiny{\textbf{Automated extractive summaries}}} \\
\hline
\tiny{\textit{NRP}} & \tiny{\textbf{Customer}} \tiny{ Can you tell me how to do Red Eye Removal in Lightroom CC? I just moved to it and don't see the Red Eye Removal tool.}\\
  & \tiny{\textbf{Agent}} { Hi Bob, here is a link to show you to use the Red eye removal in Lightroom CC.  Hi Bob, I am looping our expert team to help answer your question.}\\
\hline
\tiny{\textit{LEAD}} & \tiny{\textbf{Customer}} \tiny{  Can you tell me how to do Red Eye Removal in Lightroom CC? I just moved to it and don't see the Red Eye Removal tool.}\\
  & \tiny{\textbf{Agent}} \tiny{  Hi Bob, here is a link to show you to use the Red eye removal in Lightroom CC. Please let us know if you have any questions or need further help.}\\
\hline
\tiny{\textit{CES}} & \tiny{\textbf{Customer}} \tiny{ Can you tell me how to do Red Eye Removal in Lightroom CC? I wish a list of features missing in Lightroom CC would have been noted before I migrated my library.}\\
  & \tiny{\textbf{Agent}} \tiny{ Hi Bob, this feature is not available in Lightroom CC as of now, however you may suggest it as a feature here: [URL].  We have released Lightroom Classic CC which has all the features the old Lightroom CC 2015.12 had, you can check this article to see the differences betweem LR Classic and the new Lightroom CC: [URL].}\\

\end{tabular}
}
\caption{Two ground-truth summaries with corresponding automated summaries and QA questions}
\label{figure:summary}
\end{figure}
\endgroup

\subsection{BART models}

As a fully abstractive summarization algorithm, we used the BART model \cite{lewis2019bart} in this study. We use a lightweight variant of BART, named DistilBART provided by HuggingFace \cite{wolf2019huggingface} library\footnote{\url{https://huggingface.co/sshleifer/distilbart-cnn-12-6}}. This instance of DistilBART is fine-tuned on the extreme summarization (XSum) task, and we fine-tune this model on the \datasetName dataset. The code used for the fine-tuning is publicly available\footnote{
\url{https://github.com/huggingface/transformers/tree/master/examples/seq2seq}
}.

The hyperparameters used for training the DistilBART model are as follows:

\begin{Verbatim}[fontsize=\small]
train_batch_size=4
eval_batch_size=4
num_train_epochs=6
model_name_or_path=sshleifer/distilbart
-xsum-12-6
learning_rate=3e-5
val_check_interval=0.1
max_source_length=512
max_target_length=80
\end{Verbatim}

\section{Sample summaries with corresponding QA questions}

Figure~\ref{figure:summary} shows an example of a \datasetName{} human-generated abstractive summary along with machine-generated summaries and their corresponding QA questions. Upon acceptance of the paper, \datasetName{} release will include the set of questions that were generated as part of the human evaluation task in the Results section.

\end{document}